\documentclass[graybox]{svmult}

\usepackage{type1cm}        
\usepackage{makeidx}         
\usepackage{graphicx}        
\usepackage{multirow}                             
\usepackage{multicol}        
\usepackage[bottom]{footmisc}
\usepackage{booktabs}
\usepackage{newtxtext}       %
\usepackage[varvw]{newtxmath}       
\usepackage[colorlinks,allcolors=blue]{hyperref}
\usepackage[ruled]{algorithm2e}
\usepackage[numbers]{natbib}

\newcommand{\orcid}[1]{\href{https://orcid.org/#1}{\includegraphics[width=8pt]{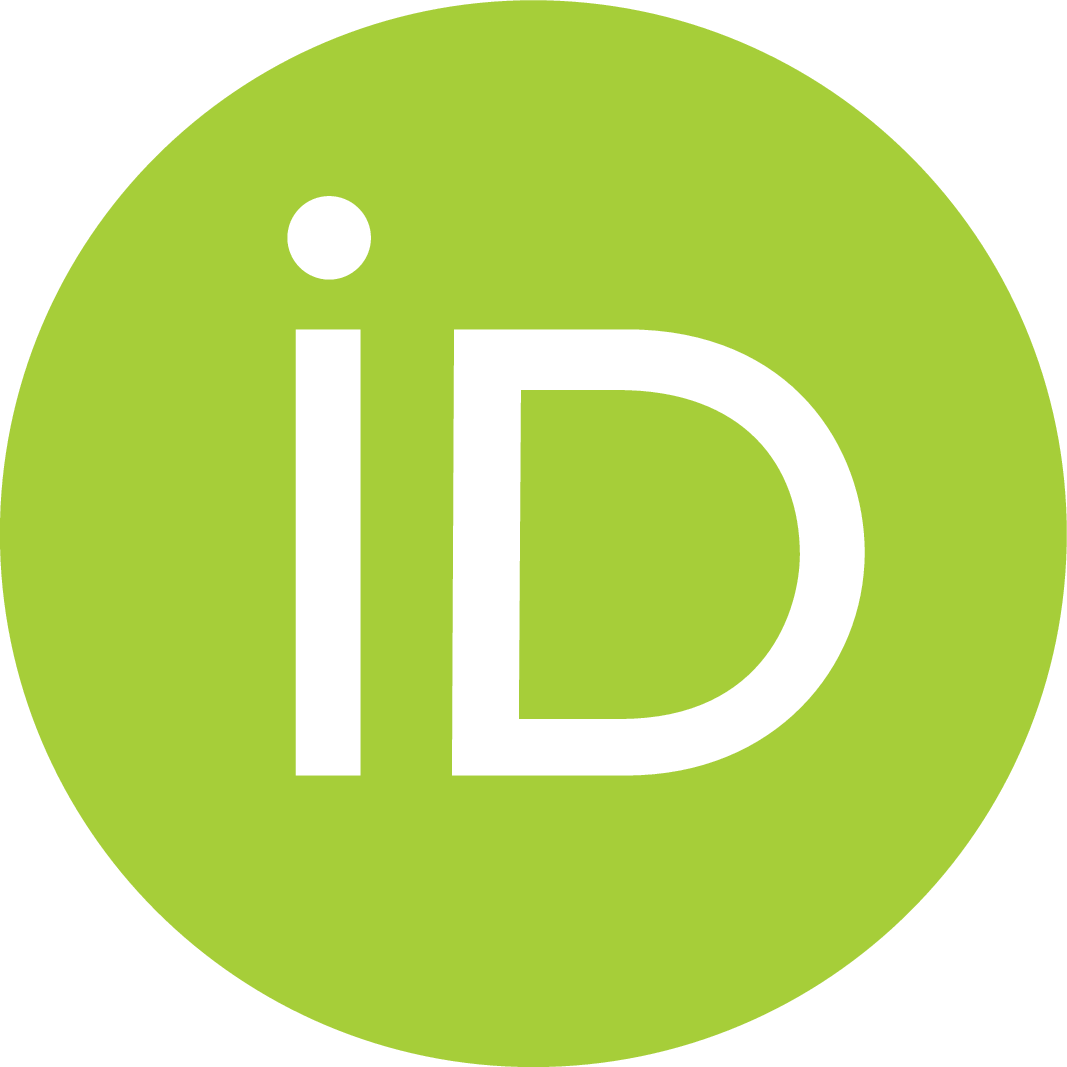}}}
\makeindex             

\begin{document}

\title*{Evolutionary Dynamic Optimization and Machine Learning}
\author{Abdennour Boulesnane\orcid{0000-0002-2272-4953}}
\authorrunning{A. Boulesnane} 
\institute{Abdennour Boulesnane \at Medicine Faculty, Salah Boubnider University Constantine 03, Constantine 25001, Algeria, \email{aboulesnane@univ-constantine3.dz}
}
%
%
\maketitle
\vspace{-2.2cm}
\abstract{     Evolutionary Computation (EC) has emerged as a powerful field of Artificial Intelligence, inspired by nature's mechanisms of gradual development. However, EC approaches often face challenges such as stagnation, diversity loss, computational complexity, population initialization, and premature convergence. To overcome these limitations, researchers have integrated learning algorithms with evolutionary techniques. This integration harnesses the valuable data generated by EC algorithms during iterative searches, providing insights into the search space and population dynamics. Similarly, the relationship between evolutionary algorithms and Machine Learning (ML) is reciprocal, as EC methods offer exceptional opportunities for optimizing complex ML tasks characterized by noisy, inaccurate, and dynamic objective functions. These hybrid techniques, known as Evolutionary Machine Learning (EML), have been applied at various stages of the ML process. EC techniques play a vital role in tasks such as data balancing, feature selection, and model training optimization. Moreover, ML tasks often require dynamic optimization, for which Evolutionary Dynamic Optimization (EDO) is valuable. This paper presents the first comprehensive exploration of reciprocal integration between EDO and ML. The study aims to stimulate interest in the evolutionary learning community and inspire innovative contributions in this domain.}

\section{Introduction}
\label{sec:1}
Evolutionary Computation (EC) is an extraordinary field of Artificial Intelligence that draws inspiration from nature's mechanisms responsible for the gradual development of intelligent organisms throughout millennia \cite{Jiang2023}. EC techniques have emerged as highly efficient and effective problem-solving methods by emulating these natural processes. These algorithms employ individuals' populations, each striving to find optimal solutions for specific challenges \cite{Miikkulainen2021}.\\
However, EC approaches face challenges that hinder their optimal performance. These obstacles often manifest as a tendency to get stuck in suboptimal solutions, diversity loss, computational complexity, population initialization, and premature convergence. Researchers have sought to integrate learning algorithms with evolutionary techniques to overcome these limitations \cite{Jiang2023}. This integration aims to address the aforementioned challenges and enhance the overall performance of EC methods. The fundamental idea behind this approach is to harness the wealth of data generated by the EC algorithm during its iterative search. This data contains valuable insights into the search space, problem characteristics, and population dynamics. By incorporating learning techniques, this data can be thoroughly analyzed and exploited to significantly improve the effectiveness of the search process \cite{Zhang2011}.

Interestingly, the relationship between evolutionary algorithms and Machine Learning (ML) goes both ways. ML tasks often involve intricate optimization problems characterized by noisy, non-continuous, non-unique, inaccurate, dynamic, and multi-optimal objective functions \cite{Qian2022}. In such complex scenarios, evolutionary computing algorithms, renowned for their versatility and stochastic search methods, offer exceptional opportunities for optimization. Consequently, several EC approaches have emerged in recent years at various stages of the ML process, ranging from pre-processing and learning to post-processing, to overcome traditional methods' limitations. These innovative hybrid techniques are collectively known as Evolutionary Machine Learning (EML) \cite{AlSahaf2019}. \\
During the pre-processing stage, utilizing EC techniques becomes valuable for tasks such as data balancing and feature selection. As we delve into the ML learning phase, EC approaches autonomously play a vital role in determining essential components of ML models, encompassing hyperparameters and architecture. Furthermore, in the post-processing stage, EC can be applied, for instance, to optimize rules or facilitate ensemble learning. Therefore, the integration of EC techniques into ML signifies a profound paradigm shift, empowering the development of ML systems that are both robust and adaptable \cite{Telikani2021}.

ML tasks are typically carried out in a real-world setting that undergoes dynamic changes, leading to the need for dynamic optimization. In such scenarios, the optimization problem's objective, constraint, or solution space can undergo variations over time. Evolutionary Dynamic Optimization (EDO) is a widely used approach that can swiftly adapt to these dynamic changes, making it valuable for practical applications in dynamic optimization. In this context, this paper introduces an innovative investigation into the convergence of EDO and ML techniques. To the best of our knowledge, this study represents a thorough investigation into the reciprocal relationship between EDO and ML. By offering up-to-date insights on using EDO approaches in ML and the integration of ML into EDO, our primary objective is to stimulate interest and inspire the evolutionary learning community, fostering the development of innovative contributions in this domain.\\

In the remainder of our paper, we introduce EDO in Section \ref{sec:2}, discussing its principles for solving optimization problems in dynamic environments. Section \ref{sec:3} provides an overview of ML. Section \ref{sec:4} focuses on applying ML to resolve dynamic optimization problems. Section \ref{sec:5} delves into utilizing EDO in ML, showcasing the synergy between these two domains. Finally, in Section \ref{sec:6}, we conclude by summarizing our investigation and suggesting future research directions.
\section{Evolutionary Dynamic Optimization}
\label{sec:2}
Dynamic optimization, also known as optimization in dynamic environments, is a highly active and extensively researched field due to its direct applicability to real-world problems. The inherent challenge lies in addressing the dynamic nature of these problems, which require finding optimal solutions within specific time constraints. These types of problems are called Dynamic Optimization Problems (DOPs) \cite{Yazdani2021}.

Formally, a DOP can be defined as the task of finding the optimal solutions $(x_{1}^{*}, x_{2}^{*},..., x_{n}^{*})$ that optimizes the time-dependent objective function $f(x,t)$, as follows:
\begin{equation}
	\label{equ:1}
	\begin{array}{l}
		\mbox{Optimize (max/min)} f(x,t) \\
		\mbox{subject to. } h_{j}(x,t)=0 \mbox{ for } j=1,2,..., u   \\
		g_{k}(x,t)\le 0 \mbox{ for } k=1,2,..., v.    \\
		\mbox{with } x\in \mathbb{R}^{n} 
	\end{array}
\end{equation}
Where  $ h_{j}(x,t) $ represents the $ j^{th} $ equality constraint and $ g_{k}(x,t) $ represents the $ k^{th} $ inequality constraint.\\
In the literature, Dynamic Multiobjective Optimization (DMO) \cite{Wang2023}, Dynamic Constrained Optimization (DCO) \cite{Hamza2022}, Robust Optimization over Time (ROOT) \cite{Yazdani2023}, and Dynamic Time-Linkage Optimization (DTO) \cite{Zhang2022} are closely related to DOPs. This latter revolves around optimizing systems that change over time, and these specific classes of problems delve into various aspects within such dynamic contexts. DMO focuses on simultaneously optimizing multiple conflicting objectives as the system evolves. CDO deals with optimization problems with constraints that must be satisfied throughout the dynamic process. ROOT emphasizes finding resilient and robust solutions against uncertainties and changes over time. DTO tackles the optimization of interdependencies and linkages between different time steps or stages of a dynamic problem. Together, these classes of problems encompass a comprehensive range of challenges and considerations when addressing dynamic optimization scenarios.\\

\begin{figure}[h]
 	\centering
 	\includegraphics[width=0.5\textwidth]{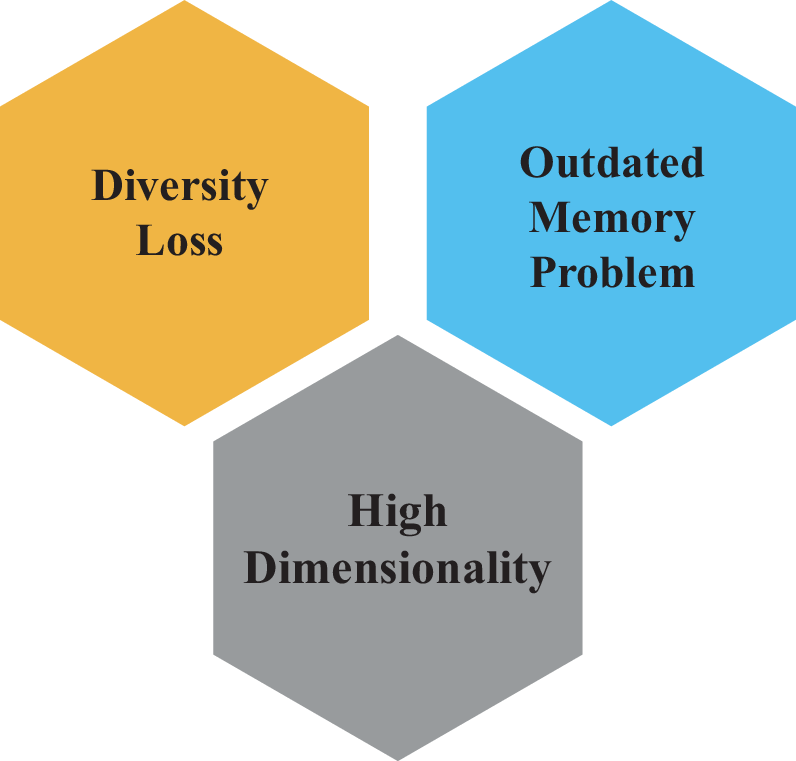}
 	\caption{Complex Challenges in the Realm of Evolutionary Dynamic Optimization.}
 	\label{figure1}
 \end{figure}

Solving DOPs requires an algorithm to find the best solution and adapt to environmental changes efficiently. However, DOPs present complex challenges, including issues like diversity loss, the persistence of outdated memory, and coping with large-scale dimensions, as depicted in Figure \ref{figure1}. Diversity loss happens when all potential solutions converge to a single area in the search space, posing difficulties in exploring new optima after environmental changes. At the same time, the outdated memory problem arises when the information accumulated during the search process becomes invalid or irrelevant after a dynamic change occurs. The integration of obsolete knowledge during the search process can yield misleading outcomes and impede the dynamic optimizer's aptitude to locate the global optimum amidst a dynamic environment. \cite{Boulesnane2022}. Furthermore, with the emergence of big data, another recent challenge is sometimes the high dimensionality of DOPs. Dealing with large-scale dimensions is a significant challenge when facing DOPs \cite{Yazdani2020}. Exploring all potential solutions becomes challenging as the problem's complexity and scale escalate, leading to an exponential growth of the search space. Large-scale dimensions often result in longer computation times and increased memory requirements, making the optimization process computationally expensive.\\
 
Effectively solving DOPs requires addressing these critical challenges. In this regard, researchers have drawn inspiration from biological evolution and natural self-organized systems in the last two decades, leading to the widespread use of Evolutionary Algorithms (EAs) and Swarm Intelligence (SI) methods \cite{Yazdani2021}. These techniques offer inherent adaptability and resilience to handle environmental changes, making them well-suited for optimizing DOPs. The field dedicated to applying EAs and similar approaches to address DOPs is known as Evolutionary Dynamic Optimization (EDO). By using the principles of EAs and SI, EDO aims to tackle the complexities of DOPs and provide efficient solutions that can adapt to dynamic environments.
\section{Machine Learning}
\label{sec:3}
\begin{figure}[h]
	\centering
	\includegraphics[height=4cm]{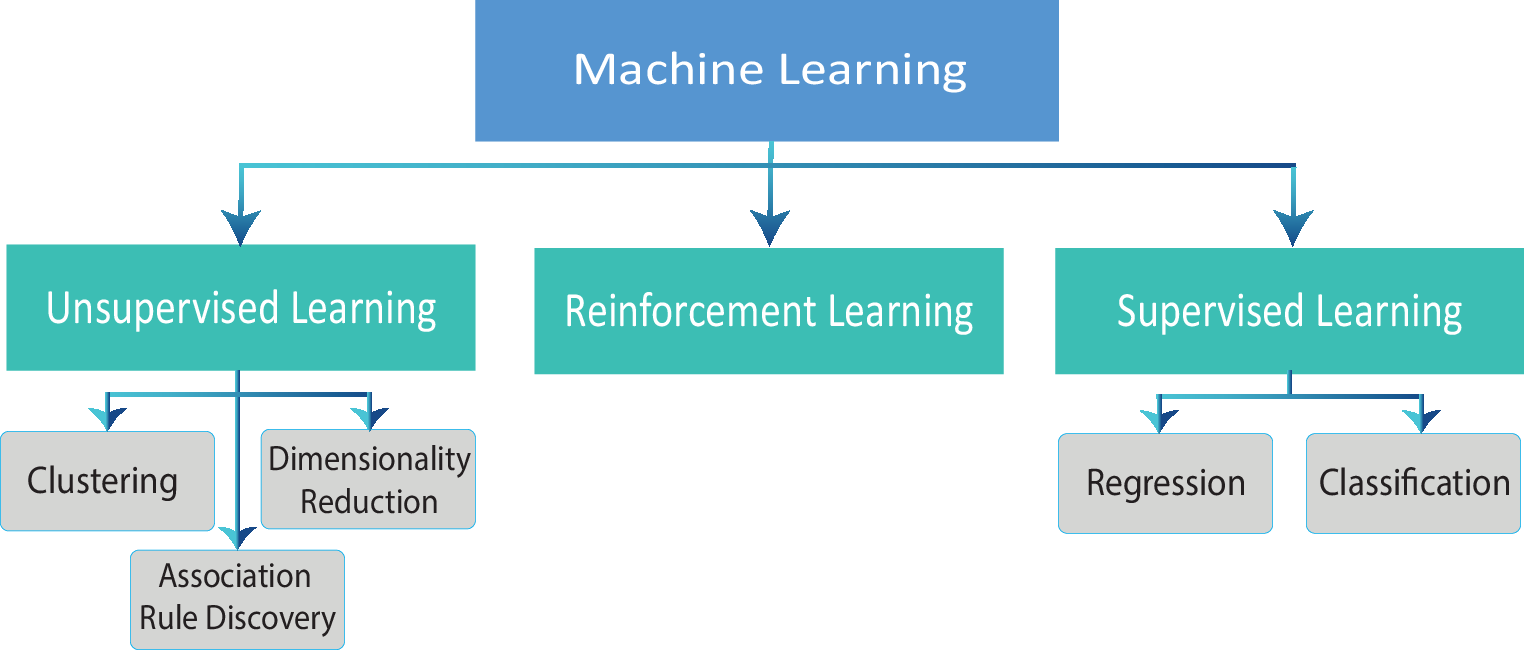}
	\caption{Taxonomy of Machine Learning.}
	\label{figure2}
\end{figure}

Artificial intelligence encompasses a wide range of techniques, and machine learning (ML) is one specific subset. ML empowers computer systems to learn and improve from data without explicit instructions \cite{Zhou2021}. It encompasses various techniques designed to enable computers to analyze and interpret vast amounts of information effectively. As shown in Figure \ref{figure2}, three primary types of ML techniques exist: Supervised, Unsupervised, and Reinforcement learning \cite{Zhang_2010}.
\begin{itemize}
	\item \textbf{Supervised learning}: involves training algorithms with labeled examples where the expected outputs are already known. The algorithm discerns underlying patterns and relationships by examining these labeled instances, effectively learning a function that maps inputs to outputs. This technique enables predictions and classifications for new, unseen data points. Examples of supervised learning algorithms include decision trees, random forests, support vector machines, and neural networks. \cite{Hastie2008}.
	\item \textbf{Unsupervised learning}: allows algorithms to learn from unlabeled data without predefined outputs. In this scenario, the algorithms independently discover hidden patterns and structures within the data without explicit guidance. Unsupervised learning utilizes clusterings and dimensionality reduction algorithms, like k-means clustering and principal component analysis (PCA), to detect patterns or resemblances within the data, enabling the exploration and extraction of insights and knowledge \cite{Alloghani2019}.
	\item \textbf{Reinforcement learning}: focuses on training software agents to learn through trial and error while interacting with an environment. The agents are given feedback in the form of rewards or penalties based on their actions, and they make choices with the goal of maximizing the total rewards they accumulate over a period of time. This technique is particularly valuable in game playing, robotics, and autonomous systems, where agents can continuously learn and adapt their behavior to achieve desired goals \cite{Gosavi2009}.
\end{itemize}	

Apart from the above paradigms, there are other valuable ML techniques. Semi-supervised learning \cite{Hady2013} combines labeled and unlabeled data to improve training efficiency. Transfer learning \cite{Weiss2016} utilizes knowledge from a related problem or domain with abundant labeled data to enhance performance on a target problem with limited training data. Multitask learning \cite{Zhang2017} trains models to tackle multiple related tasks simultaneously, exploiting shared information and differences to enhance learning efficiency and prediction accuracy. These techniques broaden the scope of ML, allowing for more effective use of data and knowledge to achieve better results in various applications. Furthermore, Deep learning is another significant ML subfield that has recently gained immense popularity \cite{LeCun2015}. It has impressively advanced the field of ML, pushing the boundaries of what computers can achieve in tasks such as image classification, object detection, machine translation, and more \cite{Valentina2019}.
\section{Machine Learning for Resolving Dynamic Optimization Problems}
\label{sec:4}
Integrating ML in optimization enables more efficient and accurate problem-solving, enhancing decision-making processes in diverse fields such as logistics, finance, healthcare, and manufacturing \cite{chelouah2022optimization}. ML algorithms can effectively learn patterns, relationships, and trends to find optimal solutions by using the accumulated data during optimization. Furthermore,  ML can be combined with traditional optimization methods to create hybrid approaches that embrace the strengths of both paradigms \cite{calvet2017learnheuristics}.\\
\begin{table}[h]
	\centering
	\caption{ML Techniques Used for Resolving DOPs and Corresponding References.}
	\label{tab1}       
	\renewcommand{\arraystretch}{1.5}%
	\begin{tabular}{llc}
		\toprule
		\textbf{ML Technique}                        & \textbf{DOP Type} & \textbf{References} \\ \midrule
		Transfer Learning                    & DMOPs    & \cite{Jiang2018tran}, \cite{Wang2023}, \cite{Jiang2021Manifold}, \cite{Zhang2023}, \cite{Jiang2021}, \cite{WANG2019}, \cite{Yao2023}, \cite{Zhang2023Elitism}, \cite{Fan2020}, \cite{Ye2022Bayesian}      \\ \midrule
		\multirow{2}{*}{Supervised Learning} & DMOPs    & \cite{Li2021},\cite{Cao2020},\cite{Jiang2018},\cite{Zhang2022Inverse}, \cite{Han2023}, \cite{Liu2022}, \cite{HU2019}, \cite{Xu2022}, \cite{Wu2023}      \\ \cline{2-3} 
		& DOPs     & \cite{Meier2018}, \cite{Meier2018b}, \cite{Liu2018}, \cite{Meier2019}, \cite{Liu2020}, \cite{Shoreh2020},  \cite{Kalita2019}      \\ \midrule
		\multirow{2}{*}{Reinforcement Learning}   & DMOPs    &      \cite{Zou2021}      \\ 
		\cline{2-3} 
		& DOPs     &     \cite{Boulesnane2021}, \cite{Talaat2022}       \\ \cline{2-3} 
		& DTPs     &    \cite{Zhang2022}        \\ \midrule
		\multirow{2}{*}{Unsupervised Learning}   & DMOPs    &    \cite{Wang2021}        \\ \cline{2-3} 
		& DOPs     &       \cite{Halder2013}, \cite{Li2009}, \cite{Yang2010}, \cite{Li2012}, \cite{Vellasques2015}, \cite{Cuevas2021}     \\ \midrule
		Deep Learning                        & DMOPs    &     \cite{Zhu2022}       \\ \midrule
		Ensemble Learning                    & DMOPs    &    \cite{Wang2020}        \\ \midrule
		Online Learning                      & DMOPs    &     \cite{Liu2021}       \\ \midrule
		Dual Learning                        & DMOPs    &    \cite{Yan2023}        \\ \bottomrule
	\end{tabular}
\end{table}

On the other hand, DOPs have attracted significant attention due to their capacity to capture the nonstationary nature inherent in real-world problems. These problems require robust algorithms to discover optimal solutions in ever-changing or uncertain environments. Given these challenges, researchers have increasingly embraced the application of ML paradigms as powerful tools to address DOPs, supplementing traditional EC approaches. By incorporating ML paradigms like Transfer Learning, Supervised Learning, Reinforcement Learning, and others (see Table \ref{tab1}), the goal is to enhance the adaptability and efficiency of algorithms. This enables them to dynamically adjust, learn and evolve their strategies in response to changing problem landscapes.
\subsection{Transfer Learning-Based}
In the literature realm, many approaches and algorithms have been put forth to tackle the complexities of dynamic multi-objective optimization problems (DMOPs) by harnessing the power of transfer learning techniques \cite{Jiang2018tran}. DMOPs encompass optimization problems with multiple conflicting objectives that evolve over time, posing a significant hurdle in effectively tracking the ever-changing set of Pareto-optimal solutions \cite{Wang2023}.\\
Transfer Learning, a method that involves leveraging past experiences and knowledge gained from previous computational processes, has garnered considerable attention due to its capacity to adapt to environmental changes and tap into valuable knowledge acquired in the past \cite{Jiang2018tran}. By employing transfer learning in DMOPs, the task of efficiently and accurately tracing the evolving Pareto-optimal fronts is greatly facilitated. \\
Various methods have been proposed to exploit past experiences and enhance the optimization process's performance, including manifold transfer learning \cite{Jiang2021Manifold,Zhang2023}, individual transfer learning \cite{Jiang2021}, regression transfer learning \cite{WANG2019}, and clustering difference-based transfer learning \cite{Yao2023}. \\
\begin{figure}[h]
	\centering
	\includegraphics[width=0.47\textheight]{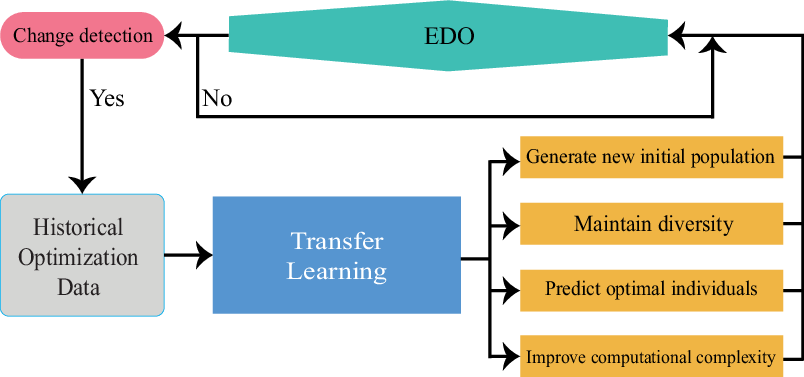}
	\caption{Synergizing Transfer Learning and EDO.}
	\label{figure3}
\end{figure}
Nonetheless, some common issues have been identified, such as the loss of population diversity and high computational consumption. To overcome these challenges, hybrid methods have emerged as a solution. These methods often integrate transfer learning with other strategies, such as elitism-based mechanisms \cite{Zhang2023Elitism}, surrogate models \cite{Fan2020}, manifold learning \cite{Zhang2023}, or Bayesian classification \cite{Ye2022Bayesian}. The goal of combining these techniques is to improve the quality of the initial population, accelerate convergence, maintain diversity, improve computational efficiency, and achieve robust prediction, as shown in Figure \ref{figure3}.

\subsection{Supervised Learning-Based}
\begin{figure}[h]
	\centering
	\includegraphics[width=0.47\textheight]{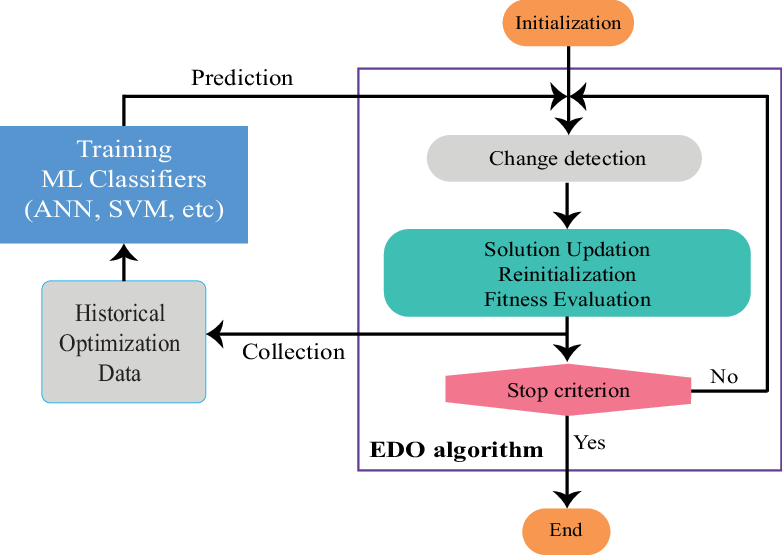}
	\caption{Employing Supervised Learning with EDO to Address DOPs.}
	\label{figure4}
\end{figure}

Recently, supervised learning methods have played a crucial role in addressing DOPs by exploiting historical data and employing predictive models like artificial neural networks (ANN) \cite{Meier2018, Meier2018b, Liu2018, Meier2019, Li2021, Liu2020, Shoreh2020}, kernel ridge regression (KRR) \cite{Liu2022}, and support vector machines (SVM) \cite{Kalita2019, Cao2020, Jiang2018, Wu2023}. As depicted in Figure \ref{figure4}, these models enable the estimation and prediction of the behaviour of optimization problems in changing environments.\\
ANNs are trained using historical data to forecast future optimal solutions or estimate the position of the next optimum. They can generate initial solutions or guide the evolutionary process towards promising regions of the solution space. By accelerating convergence and improving accuracy in tracking the optimum, ANNs contribute to enhanced performance \cite{Han2023}.\\
Additionally, researchers have developed prediction mechanisms that quickly adapt to changes in the problem environment. Strategies based on KRR, Gaussian kernel functions \cite{Liu2022}, or Inverse Gaussian processs \cite{Zhang2022Inverse} are employed to anticipate future changes and optimize evolutionary algorithms accordingly. Effective predictions play a vital role in maintaining solution quality and enhancing the search performance of the population.

SVMs are extensively utilized to tackle the challenges of DOPs. These models are trained using historical data to accurately model and predict solutions or information in dynamic scenarios. SVMs can capture linear and nonlinear correlations between past and present solutions, making them well-suited for this task. Notably, SVM-based DOPs often employ incremental learning \cite{HU2019,Xu2022}, continuously updating the SVM model with the latest optimal solutions obtained from previous periods. This iterative approach effectively incorporates knowledge from all historical optimal solutions. As a result, incremental learning facilitates real-time exploration of nonlinear correlations between solutions, enhancing the model's predictive capabilities.
\subsection{Reinforcement Learning-Based}
The utilization of reinforcement learning (RL) methods to tackle DOPs is an emerging and promising research field that warrants heightened attention. While conventional approaches in dynamic optimization have predominantly focused on EDO techniques, limited studies have explored the application of RL techniques in resolving DOPs.
\begin{figure}[h]
	\centering
	\includegraphics[width=0.47\textheight]{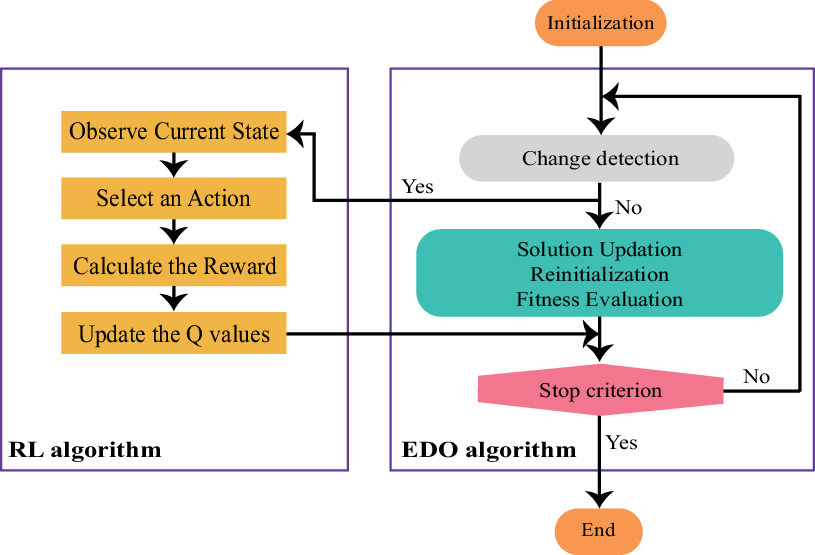}
	\caption{Illustration of the RL-DMOEA Algorithm \cite{Zou2021}.}
	\label{figure5}
\end{figure}

Among the few studies conducted, Zou et al. \cite{Zou2021} introduced a groundbreaking algorithm known as RL-DMOEA (see Figure \ref{figure5}). This reinforcement learning-based dynamic multi-objective evolutionary algorithm effectively tracks the movements of Pareto fronts over time in DMOPs. It adapts to varying degrees of environmental changes by incorporating change response mechanisms. The algorithm's effectiveness is demonstrated on CEC 2015 benchmark problems.\\
Additionally, in \cite{Zhang2022}, the authors address dynamic time-linkage optimization problems (DTPs) using a dynamic evolutionary optimization algorithm named SQL-EDO. This innovative approach combines surrogate-assisted Q-learning with evolutionary optimization to handle continuous black-box DTPs. It achieves this by extracting and predicting states, employing surrogate models for evaluating Q-values, and integrating evolutionary optimization and RL techniques for long-term decision-making. The proposed algorithm outperforms comparable algorithms and demonstrates its capability to handle different dynamic changes.\\
Furthermore, researchers in \cite{Boulesnane2021} propose an innovative Q-learning RL algorithm for DOPs, drawing inspiration from EDO techniques. This algorithm draws inspiration from EDO techniques, leading to a novel perspective on defining states, actions, and reward functions. The performance of this RL model is thoroughly assessed using a customized variant of the Moving Peaks Benchmark problem, yielding compelling results that position it competitively against state-of-the-art DOPs. To provide a clearer understanding of this algorithm, we include a pseudocode representation in Algorithm \ref{algo:1}.\\
\begin{algorithm}[h]
	\caption{Pseudo Code for the Q-learning RL Algorithm in \cite{Boulesnane2021}.}
	\label{algo:1}
	Initialization: Set initial values for the Q-values, $Q(s,a)$, for all states and actions\;
	\For{each state $s$ in the state space $S$}{
		Generate a random solution $s$\;
		Evaluate the objective function $f(s,t)$\;
	}
	\Repeat{Stopping criteria}{
		\For{each state $s$ in the state space $S$}{
			\If {$ f(s,t)> f(best_{state},t)$}{
				Update the best state: $ best_{state}$ = $s$\;
				Update the best objective value: $ f(best_{state},t)$ = $f(s,t)$\;    
			}
		}
		\For{each state $s$ in the state space $S$}{
			\tcc{Implement the $\epsilon$-greedy policy}
			\eIf {rand() < $\epsilon$}{
				Randomly select an action $a$ from the action space $A$\;
			}{
				Select the action $a$ that maximizes $Q(s, a)$: $a$ = arg $max_{{a}}(Q({s}, {a}))$\;
			}
			Execute action $a$ and observe the resulting reward $R$\;
			Compute the maximum Q-value for the next state: ${a}'$ = $max(Q(s, A))$\;
			Calculate the TD target: $TD\_target$ = $R$ + $\gamma$ * $Q(s,{a}')$\;
			Update the Q-value: $Q(s,a)$ =$Q(s,a)$ + $\alpha$*($TD\_target$ - $Q(s,a)$)\;	
		}
	}
\end{algorithm}
Moreover, in a separate study referenced as \cite{Talaat2022}, the same algorithm is successfully applied to hyperparameter optimization for Convolutional Neural Networks (CNNs), demonstrating its promising capabilities in this context as well.

By harnessing RL's capacity to learn from the environment and make decisions based on accumulated knowledge, these studies provide valuable insights into the application of RL in the realm of DOPs. Furthermore, they demonstrate promising outcomes in comparison to traditional dynamic optimization algorithms.
\subsection{Unsupervised Learning-Based}
Numerous scientific studies have employed the power of unsupervised learning techniques to tackle the intricate challenges posed by DOPs. Among the prominent methodologies employed are Clustering, Gaussian Mixture Model (GMM), and mean shift algorithms, each offering unique insights and contributions to the field.\\
Clustering strategies have been utilized effectively to navigate dynamic fitness landscapes with multiple peaks. For instance, in the paper by Halder et al. \cite{Halder2013}, the cluster-based dynamic differential evolution with an external archive algorithm is introduced. This algorithm incorporates adaptive clustering within a multi-population framework, enabling periodic information sharing among clusters. Similarly, studies in \cite{Li2009,Yang2010} present clustering particle swarm optimizers that incorporate hierarchical clustering and nearest neighbor search strategies to locate and track peaks, complemented by fast local search methods for refinement. Furthermore, Li et al. \cite{Li2012} explore hierarchical clustering in dynamic optimization and introduce a random immigrants method to reduce redundancy without relying on change detection.
\begin{figure}[h]
	\centering
	\includegraphics[width=0.47\textheight]{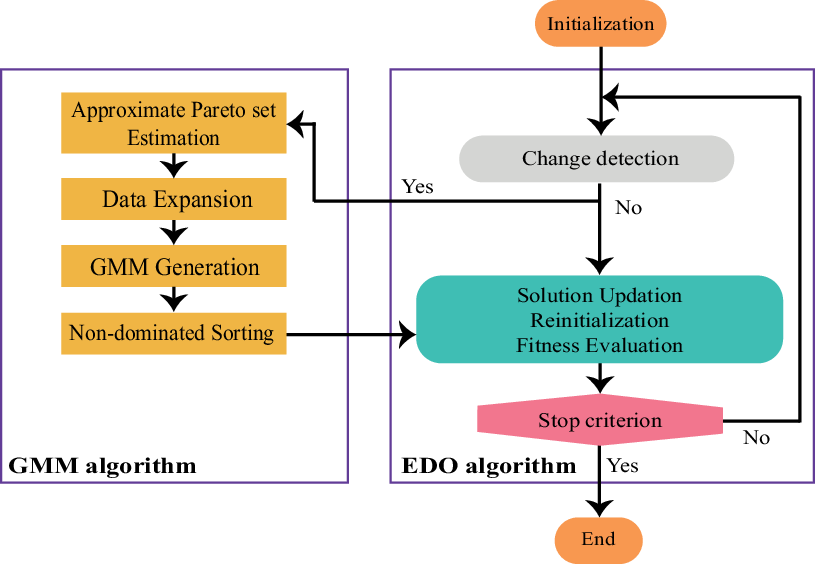}
	\caption{Flowchart Representation of the MOEA/D-GMM Algorithm \cite{Wang2021}.}
	\label{figure6}
\end{figure}

Another unsupervised learning technique is the GMM, a statistical model known for its prowess in representing intricate data patterns. GMMs play a pivotal role in enhancing the performance and effectiveness of solving DOPs. The study in \cite{Wang2021} introduces MOEA/D-GMM (see Figure \ref{figure6}), a prediction method that integrates GMMs into the framework to accurately track the changing Pareto set in DMOPs. Work in \cite{Vellasques2015} presents GMM-DPSO, a memory-based approach that efficiently optimizes streams of recurrent problems in recurrent DOPs using GMMs.

Furthermore, Cuevas et al. \cite{Cuevas2021} have made significant advancements by modifying the mean shift algorithm to effectively detect global and local optima in DOPs. The mean shift algorithm is an iterative and non-parametric process used to identify local maxima in a density function based on a set of samples. The authors enhance the algorithm by incorporating the density and fitness value of candidate solutions. This modification allows the algorithm to prioritize regions with higher fitness, which is particularly advantageous in optimization problems aiming to find the best solution.
\subsection{Other Learning-Based Models}
In addition to previous research, several alternative ML paradigms have been investigated for tackling DOPs. Although not yet widely adopted, these approaches have shown significant effectiveness and promising outcomes, highlighting the need for further exploration.

In a study conducted by Zhu et al. \cite{Zhu2022}, a method centered around deep multi-layer perceptrons was introduced. This method employed historical optimal solutions to generate an initial population for the algorithm when faced with novel environments. Another approach introduced an ensemble learning-based prediction strategy \cite{Wang2020} that integrated multiple prediction models to adapt to environmental changes more effectively. An online learning-based strategy employing Passive-Aggressive Regression was also integrated into an evolutionary algorithm to predict the Pareto optimal solution set in dynamic environments \cite{Liu2021}. Lastly, in \cite{Yan2023}, Yan et al. proposed a novel approach in their paper, which integrated decomposition-based inter-individual correlation transfer learning and dimension-wise learning. This combined method aimed to improve adaptability and expedite convergence in DMOPs.\\
These studies collectively demonstrate the potential of diverse ML techniques in addressing the challenges presented by dynamic environments, warranting further investigation and attention.
\section{Using Evolutionary Dynamic Optimization in Machine Learning}
\label{sec:5}
Optimization techniques are pivotal in ML, significantly impacting performance and efficiency. By minimizing loss functions during model training, algorithms like gradient descent accelerate convergence and improve accuracy. Hyperparameter tuning methods such as grid search and Bayesian optimization find optimal parameter combinations for enhanced model performance \cite{wu2019hyperparameter}. Optimization approaches assist in feature selection, identifying relevant subsets for improved results \cite{Ghamisi2015}. Additionally, optimization facilitates automated model architecture search, enabling the discovery of high-performing architectures \cite{Liu2023Archi}.\\

On the other hand, the utilization of EDO techniques in the realm of ML remains relatively scarce within the existing literature. However, a handful of notable works have emerged, delving into the vast potential of EDO across multiple facets of ML. These endeavours delve into the exploration and implementation of EDO methodologies in diverse ML domains, encompassing dynamic feature selection, hyperparameter tuning, model training optimization, and reinforcement learning. 

\begin{figure}[t]
	\centering
	\includegraphics[scale=0.47]{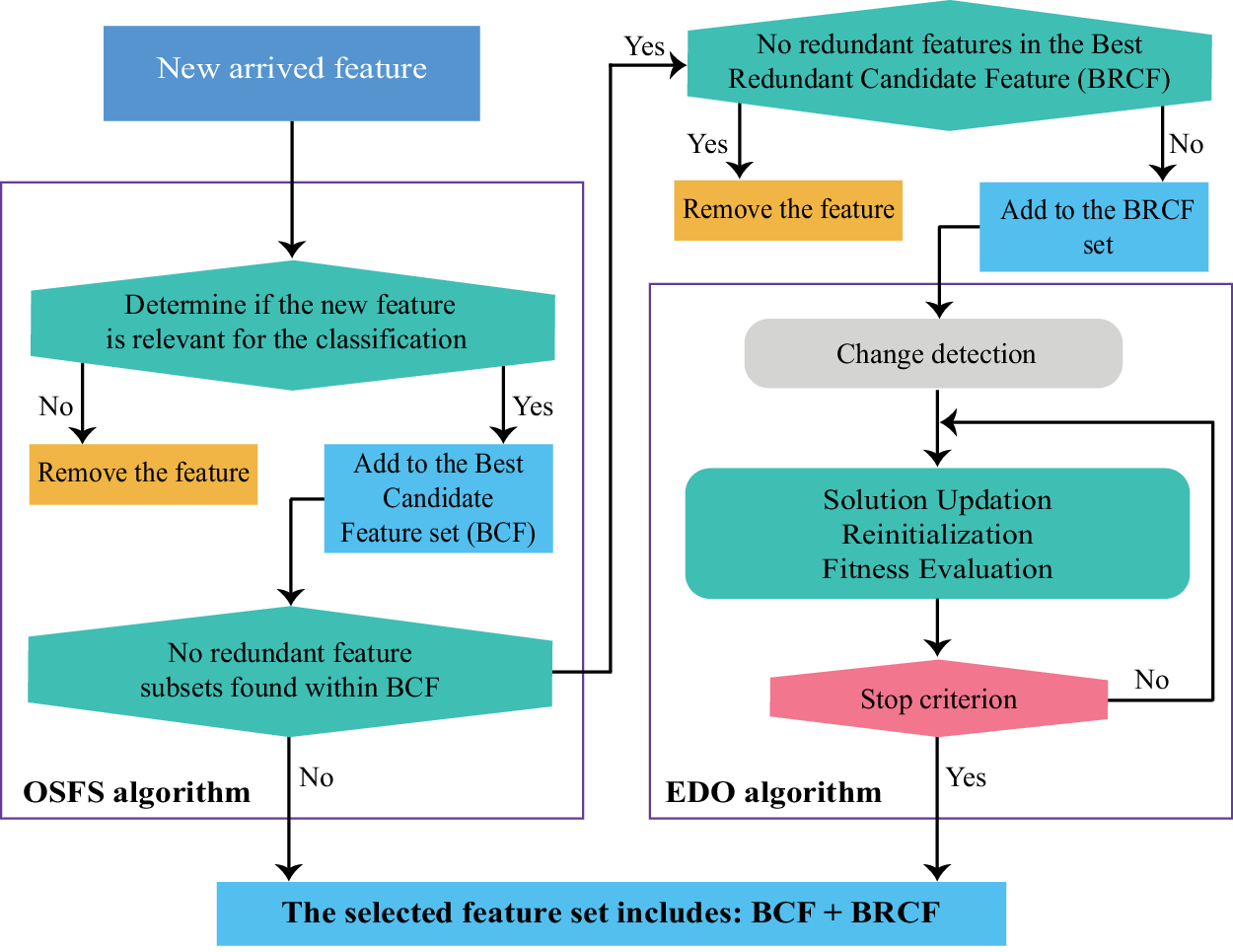}
	\caption{Streaming Feature Selection using EDO.}
	\label{figure7}
\end{figure}

The application of dynamic optimization in ML has been introduced for the first time in the literature by Boulesnane et al. \cite{Boulesnane2018}. The authors present a novel method that employs dynamic optimization to address the dynamic characteristics of streaming feature selection. The paper proposes an efficient approach for identifying relevant feature sets by combining the dynamic optimization algorithm WD2O and the Online Streaming Feature Selection (OSFS) algorithm within a hybrid model (see Figure \ref{figure7}). The goal is to find an optimal subset of attributes that enables better classification of unclassified data, considering the evolving nature of the online feature selection problem.\\
In the proposed framework, the OSFS algorithm aims to retain relevant features for classification, incorporating them with existing Best Candidate Features (BCF). Ensuring non-redundancy in BCF is vital, with irrelevant attributes being removed. However, a discarded redundant attribute could become important when interacting with new attributes. To address this, the Best Redundant Candidate Feature (BRCF) set is introduced, exclusively containing redundant attributes. The EDO algorithm then determines an optimal feature sequence using the BRCF set, independently of OSFS. Eventually, the results from both OSFS and EDO are combined to establish the final set of selected attributes.

In another similar study in \cite{Luong2021}, a novel approach has been proposed to address the challenges of classifying nonstationary data streams. The intention of the authors is to combine nonstationary stream classification and EDO by adapting the Genetic Algorithm (GA) to optimize the configuration of a streaming multi-layer ensemble. The main contribution is the introduction of SMiLE (Streaming Multi-layer Ensemble), a novel classification algorithm explicitly created for nonstationary data streams. SMiLE comprises multiple layers of diverse classifiers, and the authors additionally devise an ensemble selection approach to determine the best combination of classifiers for each layer in SMiLE. They formulate the selection process as a DOP and solve it using an adapted dynamic GA tailored for the streaming context.
\begin{table}[b]
	\centering
	\caption{Advantages and Challenges of Integrating EDO Methods in ML Tasks.}
	\label{tab2}  
	\begin{tabular}{llllc}
		\toprule
		\begin{tabular}[c]{@{}l@{}}EDO\\ Technique\end{tabular}                                          & ML Tasks                                                                                        & \multicolumn{1}{c}{Advantages}                                                                                                                                                                                                                              & \multicolumn{1}{c}{Challenges}                                                                                                                                                                                                                                            & Reference      \\ \midrule
		\begin{tabular}[c]{@{}l@{}}Dynamic\\ WD2O\end{tabular} & \begin{tabular}[c]{@{}l@{}}Streaming\\ feature\\ selection \\ and\\ classification\end{tabular} & \begin{tabular}[c]{@{}l@{}}-Enhanced Classification\\ by selecting the best\\ features, improving\\ accuracy.\\ -Considering feature\\ interactions, leading to\\ more robust models.\\ -Complement online \\ feature selection\\ algorithms.\end{tabular} & \begin{tabular}[c]{@{}l@{}}-Data drift problem.\\ -Resource Constraints.\end{tabular}   & \cite{Boulesnane2018} \\ \midrule
		\begin{tabular}[c]{@{}l@{}}Dynamic\\ GA\end{tabular}   & \begin{tabular}[c]{@{}l@{}}Streaming\\ classification\end{tabular}                              & \begin{tabular}[c]{@{}l@{}}-Enhance prediction\\ accuracy.\\ -Optimize ensemble\\ selection by dynamically\\ choosing the best\\ classifiers for each layer.\end{tabular}                                                                                  &                                                                                                                             \begin{tabular}[c]{@{}l@{}}-Data drift problem.\\ -Model Drift.\\ -Resource Constraints.\end{tabular}                                                                                                                          &  \cite{Luong2021}       \\\midrule
		\begin{tabular}[c]{@{}l@{}}Dynamic\\ AOA\end{tabular}  & \begin{tabular}[c]{@{}l@{}}Training \\ ANNs \\ for\\ classification\end{tabular}                & \begin{tabular}[c]{@{}l@{}}-Enable the training of\\ dynamic ANNs.\\ -Optimize neural network\\ parameters in real-time. \\ -Enhance classification\\ tasks.\end{tabular}                                                                                  &                                                                                                                                                                                            \begin{tabular}[c]{@{}l@{}}-Scalability challenges.\\ -Resource Constraints.\end{tabular}                                                            & \cite{Glck2023}   \\\bottomrule   
	\end{tabular}
\end{table}

The article \cite{Glck2023} introduces an advanced approach known as the Arithmetic Optimization Algorithm (AOA) to enhance the training of ANNs in dynamic environments. Although metaheuristic techniques have demonstrated effectiveness in training ANNs, their underlying assumption of static environments may not accurately capture the dynamics of real-world processes. To address this limitation, the authors of this study approach the training of ANNs as a dynamic optimization issue and introduce the AOA as a potential solution for efficiently optimizing the connection weights and biases of the ANN while accounting for concept drift. This novel method specifically focuses on improving classification tasks.

Based on EDO, these studies aim to unlock new avenues of advancement and optimization within the realm of ML, paving the way for enhanced performance and greater adaptability in the face of complex and evolving challenges.\\

Table \ref{tab2} provides a concise overview of the previously mentioned studies that employ EDO in ML tasks, highlighting both their strengths and obstacles.

\section{Conclusion}
\label{sec:6}
In this chapter, we comprehensively explore the convergence of EDO and ML techniques. EDO and ML have shown a reciprocal relationship, with each field offering valuable insights and techniques to enhance the other. We intend to inspire the evolutionary learning community to further investigate and contribute to this emerging field by highlighting the potential benefits and showcasing the promising outcomes of integrating these two domains, positing that this integration can revolutionize optimization in dynamic and complex environments.\\
Throughout this study, we have observed a significant interest in using ML, particularly transfer and supervised learning, in conjunction with dynamic multi-objective optimization compared to other approaches and problem types. However, the integration of EDO into the domain of ML remains somewhat limited, even though promising results have been achieved in prior studies. This implies that while there has been some exploration in this area, there remains significant untapped potential for further research to fully capitalize on their complementary strengths.\\
Future research in this domain should focus on developing novel EDO algorithms, refining existing ML techniques, and exploring other learning-based models to tackle the challenges of DOPs.

\bibliographystyle{unsrt} 
\bibliography{refs}

\begin{thebibliography}{10}

\bibitem{Jiang2023}
Yi~Jiang, Zhi-Hui Zhan, Kay~Chen Tan, and Jun Zhang.
\newblock Knowledge learning for evolutionary computation.
\newblock {\em {IEEE} Transactions on Evolutionary Computation}, pages 1--1,
  2023.

\bibitem{Miikkulainen2021}
Risto Miikkulainen and Stephanie Forrest.
\newblock A biological perspective on evolutionary computation.
\newblock {\em Nature Machine Intelligence}, 3(1):9--15, January 2021.

\bibitem{Zhang2011}
Jun Zhang, Zhi hui Zhan, Ying Lin, Ni~Chen, Yue jiao Gong, Jing hui Zhong,
  Henry~S.H. Chung, Yun Li, and Yu~hui Shi.
\newblock Evolutionary computation meets machine learning: A survey.
\newblock {\em {IEEE} Computational Intelligence Magazine}, 6(4):68--75,
  November 2011.

\bibitem{Qian2022}
Chao Qian.
\newblock Towards theoretically grounded evolutionary learning.
\newblock In {\em Proceedings of the Thirty-First International Joint
  Conference on Artificial Intelligence}. International Joint Conferences on
  Artificial Intelligence Organization, July 2022.

\bibitem{AlSahaf2019}
Harith Al-Sahaf, Ying Bi, Qi~Chen, Andrew Lensen, Yi~Mei, Yanan Sun, Binh Tran,
  Bing Xue, and Mengjie Zhang.
\newblock A survey on evolutionary machine learning.
\newblock {\em Journal of the Royal Society of New Zealand}, 49(2):205--228,
  April 2019.

\bibitem{Telikani2021}
Akbar Telikani, Amirhessam Tahmassebi, Wolfgang Banzhaf, and Amir~H. Gandomi.
\newblock Evolutionary machine learning: A survey.
\newblock {\em {ACM} Computing Surveys}, 54(8):1--35, October 2021.

\bibitem{Yazdani2021}
Danial Yazdani, Ran Cheng, Donya Yazdani, Jurgen Branke, Yaochu Jin, and Xin
  Yao.
\newblock A survey of evolutionary continuous dynamic optimization over two
  decades{\textemdash}part a.
\newblock {\em {IEEE} Transactions on Evolutionary Computation},
  25(4):609--629, August 2021.

\bibitem{Wang2023}
Peidi Wang and Yongjie Ma.
\newblock A dynamic multiobjective evolutionary algorithm based on fine
  prediction strategy and nondominated solutions-guided evolution.
\newblock {\em Applied Intelligence}, January 2023.

\bibitem{Hamza2022}
Noha Hamza, Saber Elsayed, Ruhul Sarker, and Daryl Essam.
\newblock Evolutionary constrained optimization with dynamic changes and
  uncertainty in the objective function.
\newblock In {\em 2022 14th International Conference on Software, Knowledge,
  Information Management and Applications ({SKIMA})}. {IEEE}, December 2022.

\bibitem{Yazdani2023}
Danial Yazdani, Donya Yazdani, J\"{u}rgen Branke, Mohammad~Nabi Omidvar,
  Amir~Hossein Gandomi, and Xin Yao.
\newblock Robust optimization over time by estimating robustness of promising
  regions.
\newblock {\em {IEEE} Transactions on Evolutionary Computation}, pages 1--1,
  2023.

\bibitem{Zhang2022}
Tuo Zhang, Handing Wang, Bo~Yuan, Yaochu Jin, and Xin Yao.
\newblock Surrogate-assisted evolutionary q-learning for black-box dynamic
  time-linkage optimization problems.
\newblock {\em {IEEE} Transactions on Evolutionary Computation}, pages 1--1,
  2022.

\bibitem{Boulesnane2022}
Abdennour Boulesnane and Souham Meshoul.
\newblock Do we need change detection for~dynamic optimization problems?: A
  survey.
\newblock In {\em Artificial Intelligence and Its Applications}, pages
  132--142. Springer International Publishing, 2022.

\bibitem{Yazdani2020}
Danial Yazdani, Mohammad~Nabi Omidvar, Jurgen Branke, Trung~Thanh Nguyen, and
  Xin Yao.
\newblock Scaling up dynamic optimization problems: A divide-and-conquer
  approach.
\newblock {\em {IEEE} Transactions on Evolutionary Computation}, 24(1):1--15,
  February 2020.

\bibitem{Zhou2021}
Zhi-Hua Zhou.
\newblock {\em Machine Learning}.
\newblock Springer Singapore, 2021.

\bibitem{Zhang_2010}
Yagang Zhang.
\newblock {\em New Advances in Machine Learning}.
\newblock IntechOpen, Rijeka, Feb 2010.

\bibitem{Hastie2008}
Trevor Hastie, Robert Tibshirani, and Jerome Friedman.
\newblock Overview of supervised learning.
\newblock In {\em The Elements of Statistical Learning}, pages 9--41. Springer
  New York, December 2008.

\bibitem{Alloghani2019}
Mohamed Alloghani, Dhiya Al-Jumeily, Jamila Mustafina, Abir Hussain, and
  Ahmed~J. Aljaaf.
\newblock A systematic review on supervised and unsupervised machine learning
  algorithms for data science.
\newblock In {\em Unsupervised and Semi-Supervised Learning}, pages 3--21.
  Springer International Publishing, September 2019.

\bibitem{Gosavi2009}
Abhijit Gosavi.
\newblock Reinforcement learning: A tutorial survey and recent advances.
\newblock {\em {INFORMS} Journal on Computing}, 21(2):178--192, May 2009.

\bibitem{Hady2013}
Mohamed Farouk~Abdel Hady and Friedhelm Schwenker.
\newblock Semi-supervised learning.
\newblock In {\em Intelligent Systems Reference Library}, pages 215--239.
  Springer Berlin Heidelberg, 2013.

\bibitem{Weiss2016}
Karl Weiss, Taghi~M. Khoshgoftaar, and DingDing Wang.
\newblock A survey of transfer learning.
\newblock {\em Journal of Big Data}, 3(1), May 2016.

\bibitem{Zhang2017}
Yu~Zhang and Qiang Yang.
\newblock An overview of multi-task learning.
\newblock {\em National Science Review}, 5(1):30--43, September 2017.

\bibitem{LeCun2015}
Yann LeCun, Yoshua Bengio, and Geoffrey Hinton.
\newblock Deep learning.
\newblock {\em Nature}, 521(7553):436--444, May 2015.

\bibitem{Valentina2019}
Valentina~Emilia Balas, Sanjiban~Sekhar Roy, Dharmendra Sharma, and Pijush
  Samui, editors.
\newblock {\em Handbook of Deep Learning Applications}.
\newblock Springer International Publishing, 2019.

\bibitem{chelouah2022optimization}
Rachid Chelouah and Patrick Siarry.
\newblock {\em Optimization and Machine Learning: Optimization for Machine
  Learning and Machine Learning for Optimization}.
\newblock John Wiley and Sons, 2022.

\bibitem{calvet2017learnheuristics}
Laura Calvet, J{\'e}sica~de Armas, David Masip, and Angel~A Juan.
\newblock Learnheuristics: hybridizing metaheuristics with machine learning for
  optimization with dynamic inputs.
\newblock {\em Open Mathematics}, 15(1):261--280, 2017.

\bibitem{Jiang2018tran}
Min Jiang, Zhongqiang Huang, Liming Qiu, Wenzhen Huang, and Gary~G. Yen.
\newblock Transfer learning-based dynamic multiobjective optimization
  algorithms.
\newblock {\em {IEEE} Transactions on Evolutionary Computation},
  22(4):501--514, August 2018.

\bibitem{Jiang2021Manifold}
Min Jiang, Zhenzhong Wang, Liming Qiu, Shihui Guo, Xing Gao, and Kay~Chen Tan.
\newblock A fast dynamic evolutionary multiobjective algorithm via manifold
  transfer learning.
\newblock {\em {IEEE} Transactions on Cybernetics}, 51(7):3417--3428, July
  2021.

\bibitem{Zhang2023}
Xi~Zhang, Guo Yu, Yaochu Jin, and Feng Qian.
\newblock An adaptive gaussian process based manifold transfer learning to
  expensive dynamic multi-objective optimization.
\newblock {\em Neurocomputing}, 538:126212, June 2023.

\bibitem{Jiang2021}
Min Jiang, Zhenzhong Wang, Shihui Guo, Xing Gao, and Kay~Chen Tan.
\newblock Individual-based transfer learning for dynamic multiobjective
  optimization.
\newblock {\em {IEEE} Transactions on Cybernetics}, 51(10):4968--4981, October
  2021.

\bibitem{WANG2019}
Zhenzhong WANG, Min JIANG, Xing GAO, Liang FENG, Weizhen HU, and Kay~Chen TAN.
\newblock Evolutionary dynamic multi-objective optimization via regression
  transfer learning.
\newblock In {\em 2019 {IEEE} Symposium Series on Computational Intelligence
  ({SSCI})}. {IEEE}, December 2019.

\bibitem{Yao2023}
Fangpei Yao and Gai-Ge Wang.
\newblock Transfer learning based on clustering difference for dynamic
  multi-objective optimization.
\newblock {\em Applied Sciences}, 13(8):4795, April 2023.

\bibitem{Zhang2023Elitism}
Xi~Zhang, Guo Yu, Yaochu Jin, and Feng Qian.
\newblock Elitism-based transfer learning and diversity maintenance for dynamic
  multi-objective optimization.
\newblock {\em Information Sciences}, 636:118927, July 2023.

\bibitem{Fan2020}
Xuezhou Fan, Ke~Li, and Kay~Chen Tan.
\newblock Surrogate assisted evolutionary algorithm based on transfer learning
  for dynamic expensive multi-objective optimisation problems.
\newblock In {\em 2020 {IEEE} Congress on Evolutionary Computation ({CEC})}.
  {IEEE}, July 2020.

\bibitem{Ye2022Bayesian}
Yulong Ye, Lingjie Li, Qiuzhen Lin, Ka-Chun Wong, Jianqiang Li, and Zhong Ming.
\newblock Knowledge guided bayesian classification for dynamic multi-objective
  optimization.
\newblock {\em Knowledge-Based Systems}, 250:109173, August 2022.

\bibitem{Li2021}
Sanyi Li, Shengxiang Yang, Yanfeng Wang, Weichao Yue, and Junfei Qiao.
\newblock A modular neural network-based population prediction strategy for
  evolutionary dynamic multi-objective optimization.
\newblock {\em Swarm and Evolutionary Computation}, 62:100829, April 2021.

\bibitem{Cao2020}
Leilei Cao, Lihong Xu, Erik~D. Goodman, Chunteng Bao, and Shuwei Zhu.
\newblock Evolutionary dynamic multiobjective optimization assisted by a
  support vector regression predictor.
\newblock {\em {IEEE} Transactions on Evolutionary Computation},
  24(2):305--319, April 2020.

\bibitem{Jiang2018}
Min Jiang, Weizhen Hu, Liming Qiu, Minghui Shi, and Kay~Chen Tan.
\newblock Solving dynamic multi-objective optimization problems via support
  vector machine.
\newblock In {\em 2018 Tenth International Conference on Advanced Computational
  Intelligence ({ICACI})}. {IEEE}, March 2018.

\bibitem{Zhang2022Inverse}
Huan Zhang, Jinliang Ding, Min Jiang, Kay~Chen Tan, and Tianyou Chai.
\newblock Inverse gaussian process modeling for evolutionary dynamic
  multiobjective optimization.
\newblock {\em {IEEE} Transactions on Cybernetics}, 52(10):11240--11253,
  October 2022.

\bibitem{Han2023}
Honggui Han, Yucheng Liu, Linlin Zhang, Hongxu Liu, Hongyan Yang, and Junfei
  Qiao.
\newblock Knowledge reconstruction for dynamic multi-objective particle swarm
  optimization using fuzzy neural network.
\newblock {\em International Journal of Fuzzy Systems}, March 2023.

\bibitem{Liu2022}
Min Liu, Diankun Chen, Qiongbing Zhang, Yizhi Liu, and Yijiang Zhao.
\newblock A dynamic multi-objective evolutionary algorithm assisted by kernel
  ridge regression.
\newblock In {\em Advances in Natural Computation, Fuzzy Systems and Knowledge
  Discovery}, pages 128--136. Springer International Publishing, 2022.

\bibitem{HU2019}
Weizhen HU, Min JIANG, Xing Gao, Kay~Chen TAN, and Yiu ming Cheung.
\newblock Solving dynamic multi-objective optimization problems using
  incremental support vector machine.
\newblock In {\em 2019 {IEEE} Congress on Evolutionary Computation ({CEC})}.
  {IEEE}, June 2019.

\bibitem{Xu2022}
Dejun Xu, Min Jiang, Weizhen Hu, Shaozi Li, Renhu Pan, and Gary~G. Yen.
\newblock An online prediction approach based on incremental support vector
  machine for dynamic multiobjective optimization.
\newblock {\em {IEEE} Transactions on Evolutionary Computation},
  26(4):690--703, August 2022.

\bibitem{Wu2023}
Xunfeng Wu, Qiuzhen Lin, Wu~Lin, Yulong Ye, Qingling Zhu, and Victor~C.M.
  Leung.
\newblock A kriging model-based evolutionary algorithm with support vector
  machine for dynamic multimodal optimization.
\newblock {\em Engineering Applications of Artificial Intelligence},
  122:106039, June 2023.

\bibitem{Meier2018}
Almuth Meier and Oliver Kramer.
\newblock Prediction with recurrent neural networks in evolutionary dynamic
  optimization.
\newblock In {\em Applications of Evolutionary Computation}, pages 848--863.
  Springer International Publishing, 2018.

\bibitem{Meier2018b}
Almuth Meier and Oliver Kramer.
\newblock Recurrent neural network-predictions for {PSO} in dynamic
  optimization.
\newblock In {\em Proceedings of the Genetic and Evolutionary Computation
  Conference}. {ACM}, July 2018.

\bibitem{Liu2018}
Xiao-Fang Liu, Zhi-Hui Zhan, and Jun Zhang.
\newblock Neural network for change direction prediction in dynamic
  optimization.
\newblock {\em {IEEE} Access}, 6:72649--72662, 2018.

\bibitem{Meier2019}
Almuth Meier and Oliver Kramer.
\newblock Predictive uncertainty estimation with temporal convolutional
  networks for dynamic evolutionary optimization.
\newblock In {\em Lecture Notes in Computer Science}, pages 409--421. Springer
  International Publishing, 2019.

\bibitem{Liu2020}
Xiao-Fang Liu, Zhi-Hui Zhan, Tian-Long Gu, Sam Kwong, Zhenyu Lu, Henry
  Been-Lirn Duh, and Jun Zhang.
\newblock Neural network-based information transfer for dynamic optimization.
\newblock {\em {IEEE} Transactions on Neural Networks and Learning Systems},
  31(5):1557--1570, May 2020.

\bibitem{Shoreh2020}
Maryam~Hasani Shoreh, Renato~Hermoza Aragones, and Frank Neumann.
\newblock Using neural networks and diversifying differential evolution for
  dynamic optimisation.
\newblock In {\em 2020 {IEEE} Symposium Series on Computational Intelligence
  ({SSCI})}. {IEEE}, December 2020.

\bibitem{Kalita2019}
Dhruba~Jyoti Kalita and Shailendra Singh.
\newblock {SVM} hyper-parameters optimization using quantized multi-{PSO} in
  dynamic environment.
\newblock {\em Soft Computing}, 24(2):1225--1241, April 2019.

\bibitem{Zou2021}
Fei Zou, Gary~G. Yen, Lixin Tang, and Chunfeng Wang.
\newblock A reinforcement learning approach for dynamic multi-objective
  optimization.
\newblock {\em Information Sciences}, 546:815--834, February 2021.

\bibitem{Boulesnane2021}
Abdennour Boulesnane and Souham Meshoul.
\newblock Reinforcement learning for dynamic optimization problems.
\newblock In {\em Proceedings of the Genetic and Evolutionary Computation
  Conference Companion}. {ACM}, July 2021.

\bibitem{Talaat2022}
Fatma~M. Talaat and Samah~A. Gamel.
\newblock {RL} based hyper-parameters optimization algorithm ({ROA}) for
  convolutional neural network.
\newblock {\em Journal of Ambient Intelligence and Humanized Computing}, March
  2022.

\bibitem{Wang2021}
Feng Wang, Fanshu Liao, Yixuan Li, and Hui Wang.
\newblock A new prediction strategy for dynamic multi-objective optimization
  using gaussian mixture model.
\newblock {\em Information Sciences}, 580:331--351, November 2021.

\bibitem{Halder2013}
U.~Halder, S.~Das, and D.~Maity.
\newblock A cluster-based differential evolution algorithm with external
  archive for optimization in dynamic environments.
\newblock {\em {IEEE} Transactions on Cybernetics}, 43(3):881--897, June 2013.

\bibitem{Li2009}
Changhe Li and Shengxiang Yang.
\newblock A clustering particle swarm optimizer for dynamic optimization.
\newblock In {\em 2009 {IEEE} Congress on Evolutionary Computation}. {IEEE},
  May 2009.

\bibitem{Yang2010}
Shengxiang Yang and Changhe Li.
\newblock A clustering particle swarm optimizer for locating and tracking
  multiple optima in dynamic environments.
\newblock {\em {IEEE} Transactions on Evolutionary Computation},
  14(6):959--974, December 2010.

\bibitem{Li2012}
Changhe Li and Shengxiang Yang.
\newblock A general framework of multipopulation methods with clustering in
  undetectable dynamic environments.
\newblock {\em {IEEE} Transactions on Evolutionary Computation},
  16(4):556--577, August 2012.

\bibitem{Vellasques2015}
Eduardo Vellasques, Robert Sabourin, and Eric Granger.
\newblock A dual-purpose memory approach for dynamic particle swarm
  optimization of recurrent problems.
\newblock In {\em Recent Advances in Computational Intelligence in Defense and
  Security}, pages 367--389. Springer International Publishing, December 2015.

\bibitem{Cuevas2021}
Erik Cuevas, Jorge G{\'{a}}lvez, Miguel Toski, and Karla Avila.
\newblock Evolutionary-mean shift algorithm for dynamic multimodal function
  optimization.
\newblock {\em Applied Soft Computing}, 113:107880, December 2021.

\bibitem{Zhu2022}
Zhen Zhu, Yanpeng Yang, Dongqing Wang, Xiang Tian, Long Chen, Xiaodong Sun, and
  Yingfeng Cai.
\newblock Deep multi-layer perceptron-based evolutionary algorithm for dynamic
  multiobjective optimization.
\newblock {\em Complex {\&} Intelligent Systems}, 8(6):5249--5264, May 2022.

\bibitem{Wang2020}
Feng Wang, Yixuan Li, Fanshu Liao, and Hongyang Yan.
\newblock An ensemble learning based prediction strategy for dynamic
  multi-objective optimization.
\newblock {\em Applied Soft Computing}, 96:106592, November 2020.

\bibitem{Liu2021}
Min Liu, Diankun Chen, Qiongbing Zhang, and Lei Jiang.
\newblock An online machine learning-based prediction strategy for dynamic
  evolutionary multi-objective optimization.
\newblock In {\em Lecture Notes in Computer Science}, pages 193--204. Springer
  International Publishing, 2021.

\bibitem{Yan2023}
Li~Yan, Wenlong Qi, Jing Liang, Boyang Qu, Kunjie Yu, Caitong Yue, and Xuzhao
  Chai.
\newblock Inter-individual correlation and dimension based dual learning for
  dynamic multi-objective optimization.
\newblock {\em {IEEE} Transactions on Evolutionary Computation}, pages 1--1,
  2023.

\bibitem{wu2019hyperparameter}
Jia Wu, Xiu-Yun Chen, Hao Zhang, Li-Dong Xiong, Hang Lei, and Si-Hao Deng.
\newblock Hyperparameter optimization for machine learning models based on
  bayesian optimization.
\newblock {\em Journal of Electronic Science and Technology}, 17(1):26--40,
  2019.

\bibitem{Ghamisi2015}
Pedram Ghamisi and Jon~Atli Benediktsson.
\newblock Feature selection based on hybridization of genetic algorithm and
  particle swarm optimization.
\newblock {\em {IEEE} Geoscience and Remote Sensing Letters}, 12(2):309--313,
  February 2015.

\bibitem{Liu2023Archi}
Yuqiao Liu, Yanan Sun, Bing Xue, Mengjie Zhang, Gary~G. Yen, and Kay~Chen Tan.
\newblock A survey on evolutionary neural architecture search.
\newblock {\em {IEEE} Transactions on Neural Networks and Learning Systems},
  34(2):550--570, February 2023.

\bibitem{Boulesnane2018}
Abdennour Boulesnane and Souham Meshoul.
\newblock Effective streaming evolutionary feature selection using dynamic
  optimization.
\newblock In {\em Computational Intelligence and Its Applications}, pages
  329--340. Springer International Publishing, 2018.

\bibitem{Luong2021}
Anh~Vu Luong, Tien~Thanh Nguyen, and Alan Wee-Chung Liew.
\newblock Streaming multi-layer ensemble selection using dynamic genetic
  algorithm.
\newblock In {\em 2021 Digital Image Computing: Techniques and Applications
  ({DICTA})}. {IEEE}, November 2021.

\bibitem{Glck2023}
İlker Gölcük, Fehmi~Burcin Ozsoydan, and Esra~Duygu Durmaz.
\newblock An improved arithmetic optimization algorithm for training
  feedforward neural networks under dynamic environments.
\newblock {\em Knowledge-Based Systems}, 263:110274, March 2023.

\end{thebibliography}
\end{document}